\documentclass{article}

\PassOptionsToPackage{numbers, compress}{natbib}
\usepackage[preprint]{neurips_2026}

\usepackage[utf8]{inputenc} 
\usepackage[T1]{fontenc}    
\usepackage{hyperref}       
\usepackage{url}            
\usepackage{booktabs}       
\usepackage{amsfonts}       
\usepackage{nicefrac}       
\usepackage{microtype}      
\usepackage{xcolor}         
\usepackage{graphicx}
\usepackage{amsmath}
\usepackage{multirow}
\usepackage{multicol}
\usepackage[table]{xcolor}
\usepackage{float}

\title{MedFlowSeg: Flow Matching for Medical Image Segmentation with Frequency-Aware Attention}

\author{
Zhi Chen$^{1}$\thanks{Equal contribution.} \quad
Runze Hu$^{2}$\footnotemark[1] \quad
Le Zhang$^{1}$\thanks{Corresponding author.} \\
$^1$School of Engineering, University of Birmingham \\
$^2$School of Public Health, Peking University \\
{\small
\texttt{\{zxc561@student.bham.ac.uk, 2110306221@stu.pku.edu.cn, l.zhang.16@bham.ac.uk\}}
}
}

\begin{document}

\maketitle

\begin{abstract}
Flow matching has recently emerged as a principled framework for learning continuous-time transport maps, enabling efficient ODE-based sampling without relying on stochastic diffusion processes. While generative modeling has shown promise for medical image segmentation, particularly in capturing uncertainty and complex anatomical variability, existing approaches are predominantly based on diffusion models, which require iterative sampling and incur substantial computational overhead.
In this work, we propose \textbf{MedFlowSeg}, a conditional flow matching framework that formulates medical image segmentation as learning a time-dependent vector field that transports a simple prior distribution to the target segmentation distribution. Compared to diffusion-based methods, our formulation enables more efficient inference through solving an ordinary differential equation, while preserving the flexibility of generative modeling.
To effectively incorporate conditional information, we introduce a dual-conditioning mechanism. Specifically, we propose a \textit{Dual-Branch Spatial Attention} (DB-SA) module to inject multi-frequency structural priors, and a \textit{Frequency-Aware Attention} (FA-Attention) module to model interactions between spatial and spectral representations via discrepancy-aware fusion and time-dependent modulation. These components improve the alignment between noisy intermediate states and clean semantic features, leading to better structural consistency and boundary delineation.
We conduct extensive experiments across multiple medical imaging modalities, where MedFlowSeg consistently outperforms prior state-of-the-art (SOTA) baselines, including diffusion-based and flow-based methods.
\textit{Code is available at} \url{https://github.com/yyxl123/MedFlowSeg}.
\end{abstract}

\section{Introduction}

Medical image segmentation aims to partition medical images into meaningful anatomical regions, which plays a crucial role in many clinical applications such as diagnosis, treatment planning, and image-guided interventions. In recent years, deep learning-based approaches have achieved remarkable success in this field, including classical convolutional neural networks (CNNs) such as U-Net and its variants~\cite{zhou2018unet++, ronneberger2015unet}, as well as transformer-based architectures that enhance global contextual modeling~\cite{chen2021transunet, wu2022seatrans, ji2021learning}. More recently, generative models, particularly diffusion probabilistic models (DPMs)~\cite{ho2020denoising}, have shown strong potential for medical image segmentation by modeling structural uncertainty and producing diverse predictions~\cite{wu2024medsegdiff,wu2024medsegdiffv2}. 

Despite their success, diffusion-based segmentation methods suffer from several limitations. First, they rely on iterative denoising processes with a fixed number of sampling steps, resulting in limited flexibility during inference. Second, these methods often struggle to effectively integrate conditional features with the denoising process, leading to suboptimal alignment between noisy intermediate representations and clean semantic priors. As a result, their performance is typically sensitive to specific task settings, achieving strong results on certain benchmarks while exhibiting limited generalization across different modalities and anatomical structures. Although recent works attempt to improve conditional integration, they still lack an effective mechanism to address this issue.

Motivated by these challenges, we explore an alternative generative paradigm for medical image segmentation based on flow matching. Compared with diffusion models, flow matching directly learns a continuous deterministic velocity field~\cite{lipman2022flow,liu2022flow}, enabling more efficient generation. More importantly, the sampling process is inherently flexible and can be adaptively adjusted according to task requirements, allowing a better trade-off between efficiency and accuracy. Recent studies~\cite{bogensperger2025flowsdfijcv} have shown that adopting UNet-style backbones within the flow matching framework can already achieve competitive performance in medical image segmentation. However, existing flow-based methods still struggle to effectively leverage conditional information, particularly when aligning noisy intermediate representations with clean anatomical priors, and are often tailored to specific tasks~\cite{bogensperger2025flowsdfijcv,ngoc2025latentfm}, limiting their generalization across diverse modalities and segmentation scenarios.

To address these issues, we propose \textbf{MedFlowSeg}, a novel flow matching framework for medical image segmentation. 
The core idea is to explicitly model the discrepancy between noisy intermediate flow states and clean conditional priors.
To this end, we introduce a dual-conditioning architecture that decomposes conditional guidance into structural alignment and semantic alignment. 
For structural alignment, MedFlowSeg injects multi-frequency anatomical priors into shallow flow representations, enabling stable guidance for both coarse structures and fine boundaries. 
For semantic alignment, we model cross-domain interactions in both spatial and spectral spaces, allowing discrepancy-aware feature fusion and time-dependent modulation between noisy flow states and clean semantic features.


\textbf{Our main contributions can be summarized as:}
\textbf{(1)} We are the first to integrate attention mechanisms into a flow-based generative model for general medical image segmentation, improving the effectiveness of conditional information modeling.
\textbf{(2)} We propose a unified flow matching based framework for general medical image segmentation, enabling consistent modeling across diverse tasks and imaging modalities.
\textbf{(3)} We design a dual-conditioning architecture, including a Dual-Branch Spatial Attention (DB-SA) module for multi-frequency structural guidance and a Frequency-Aware Attention (FA-Attention) mechanism for discrepancy-aware semantic alignment across spatial and spectral domains.
\textbf{(4)} Extensive experiments on multiple benchmarks across different modalities demonstrate that MedFlowSeg consistently achieves state-of-the-art performance over CNN-based, transformer-based, diffusion-based, and flow-based methods.

\section{Related Work}
\label{gen_inst}

\textbf{Deterministic Models for Medical Segmentation}
U-Net and its variants have long been the dominant architectures for medical image segmentation, owing to their effective encoder--decoder design and strong ability to preserve spatial details~\cite{ronneberger2015unet}. With the rise of vision transformers, recent works further enhance global context modeling by incorporating self-attention mechanisms, such as TransUNet, Swin-UNet, Swin-UNETR, and DS-TransUNet~\cite{chen2021transunet,cao2021swinunet,hatamizadeh2022swinunetr,lin2022dstransunet}. 
However, medical imaging data often exhibit substantial variations in acquisition devices, noise levels, and pathological characteristics, resulting in a complex data distribution. Consequently, a single deterministic prediction may fail to capture the full range of plausible interpretations. Although these models achieve strong performance on specific benchmarks, their generalization across datasets and modalities remains limited. Moreover, the lack of explicit uncertainty modeling makes them less effective in handling irregular anatomical structures and ambiguous boundaries, leading to suboptimal results in challenging cases.

\textbf{Generative Models for Medical Segmentation}
Recently, generative models have shown promising potential for medical image segmentation by reformulating it as a conditional generation problem. Diffusion-based methods, such as EnsemDiff~\cite{wolleb2021diffusion}, MedSegDiff~\cite{wu2024medsegdiff}, and MedSegDiff-V2~\cite{wu2024medsegdiffv2}, leverage stochastic sampling to model uncertainty and improve predictions on ambiguous boundaries, but rely on fixed iterative denoising, limiting inference flexibility. Flow matching provides a more efficient alternative by learning a continuous velocity field for efficient ODE-based generation with deterministic trajectories given an initial condition, with recent works such as FlowSDF~\cite{bogensperger2025flowsdfijcv} and LatentFM~\cite{ngoc2025latentfm} exploring this direction. However, existing methods still adopt simple conditioning strategies and fail to effectively bridge the gap between noisy intermediate states and clean semantic priors, limiting their generalization. Therefore, a unified flow matching framework for general medical image segmentation is needed to enable consistent modeling across diverse tasks and modalities.

\section{Methodology}
\label{3}

In this section, we present the details of the proposed MedFlowSeg framework. We first briefly review the formulation of conditional flow matching. Then, we provide an overall pipeline of the architecture, highlighting how conditional features are injected into the flow network. Finally, we elaborate on our two core conditioning mechanisms: the Dual-Branch Spatial Attention (DB-SA) for structural condition, and the Frequency-Aware Attention (FA-Attention) for semantic condition.

\subsection{Preliminaries}

\begin{figure}[t]
    \centering
    \includegraphics[width=0.85\linewidth]{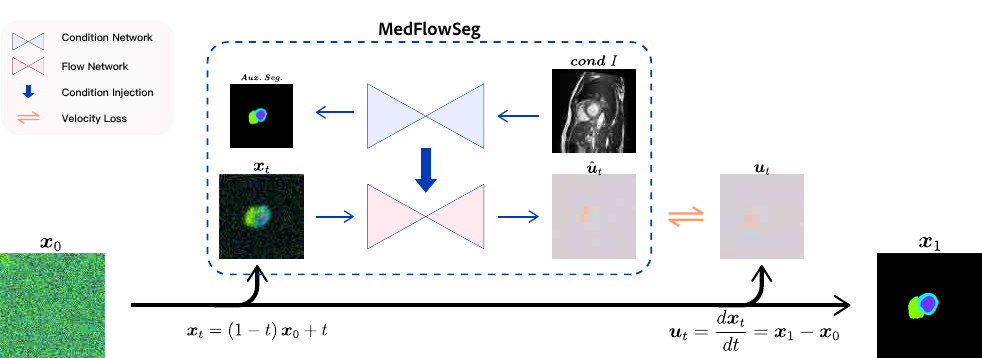}
    \caption{Overview of the proposed MedFlowSeg pipeline. The generative process is defined as a continuous trajectory from a random noise prior $x_0$ to the final segmentation label $x_1$. At a given time step $t$, the intermediate state $x_t$ is generated. The raw medical image $I$ is processed by a Condition Network to provide structural and semantic guidance to the Flow Network. The model takes $x_t$, $I$, and $t$ as inputs to predict the image-guided velocity $\hat{u}_t$. During training, the model is optimized by minimizing the velocity loss between $\hat{u}_t$ and the target velocity $u_t = x_1 - x_0$, together with an auxiliary segmentation loss applied on the condition branch.}
    \label{main}
\end{figure}

We have designed our model based on the conditional flow matching framework \cite{lipman2022flow,liu2022flow}. Unlike conventional diffusion models that reverse a multi-step noise corruption process, flow matching generative models directly learn a continuous deterministic velocity field. As illustrated by the overall pipeline in Figure~\ref{main}, MedFlowSeg establishes an end-to-end generation procedure. It defines a continuous trajectory starting from a random noise sample $x_0 \sim \mathcal{N}(0,1)$ and progressively evolving toward the final segmentation label $x_1$, structurally guided by the raw image condition $I$.
This probability path and its target velocity $u_t$ at time step $t \in [0,1]$ can be formulated as follows:
\begin{equation}
\begin{gathered}
    x_t = (1-t)x_0 + t \cdot x_1,\\
    u_t = \frac{dx_t}{dt} = x_1 - x_0,
\end{gathered}
\end{equation}
where $x_t$ represents the intermediate state along the path, and $u_t$ denotes the constant target velocity pointing directly from the noise prior $x_0$ to the clean segmentation $x_1$.
Within this pipeline, we utilize a neural network parameterized by $\theta$ to estimate the velocity. To achieve precise segmentation, conditional features extracted from the raw medical image $I$ are injected into the flow network:
\begin{equation}
    \hat{u}_t = F_\theta(x_t, I, t),
\end{equation}
where the network $F_\theta$ takes the intermediate state $x_t$, the condition $I$, and the time step $t$ as inputs to predict the image-guided velocity $\hat{u}_t$.
During training, the predicted velocity $\hat{u}_t$ is constrained to match the target velocity $u_t$. Notably, instead of the conventional mean squared error, our pipeline specifically minimizes the expected $L_1$ error. This design choice is employed to robustly penalize velocity estimation errors and better preserve sharp structural boundaries in the segmentation:
\begin{equation}
\label{loss::vel}
    \mathcal{L}_{\mathrm{vel}} = \mathbb{E}_{x_0, x_1, t} \left[ \left\| \hat{u}_t - u_t \right\|_1 \right].
\end{equation}

In addition to the velocity supervision, the conditional network is equipped with an auxiliary segmentation head to provide direct structural guidance. This auxiliary objective is denoted as $\mathcal{L}_{\mathrm{aux}}$, which will be detailed in Section~\ref{3.2}.
At inference time, starting from a random initialization $x_0 \sim \mathcal{N}(0,1)$, the pipeline recovers the clear mask step by step by solving an Ordinary Differential Equation (ODE) along the predicted velocity field, ultimately yielding the final segmentation result $x_1$.

\subsection{Overall Architecture}
\label{3.2}

\begin{figure}[t]
    \centering
    \includegraphics[width=0.85\linewidth]{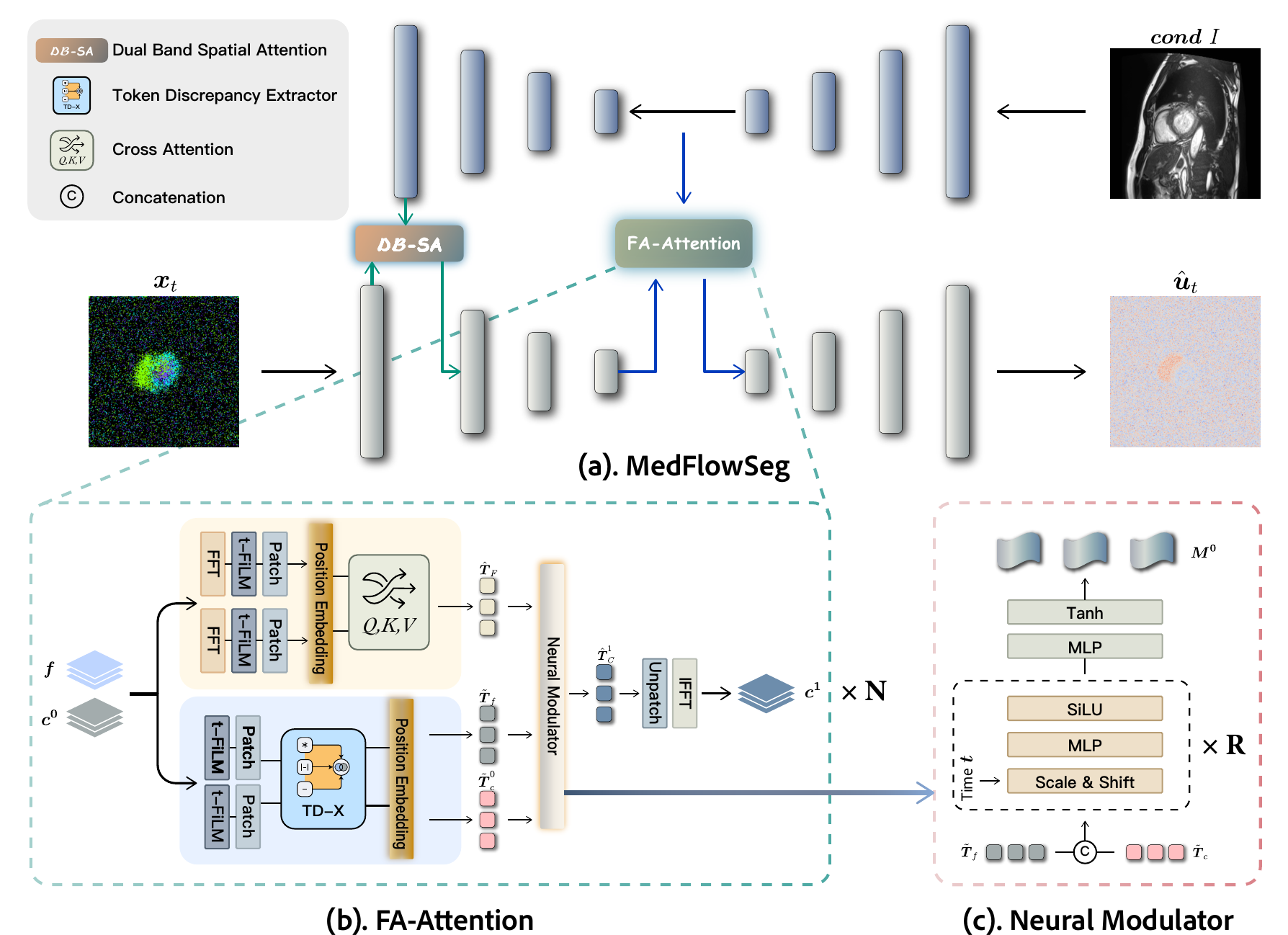}
    \caption{An illustration of MedFlowSeg, which starts from (a) an overview of the two-stream architecture, and continues with zoomed-in diagrams of individual modules, including (b) FA-Attention, and (c) Neural Modulator.}
    \label{framework}
\end{figure}

The overall architecture of MedFlowSeg is shown in Figure~\ref{framework}(a). At each time step $t$ of the flow matching trajectory, the intermediate state $x_t$ is fed into a UNet-style Flow Network, while the raw medical image $I$ is processed by a parallel standard UNet, termed the Condition Network.

To bridge the two streams, we introduce two complementary conditioning mechanisms, namely a structural condition and a semantic condition. The structural condition is imposed at the shallow stage of the Flow Network through Dual-Branch Spatial Attention (DB-SA). DB-SA takes the final high-resolution decoder feature of the Condition Network as input, which is transformed and injected into the first encoder stage of the Flow Network. By decomposing this structural prior into low-frequency and high-frequency components, DB-SA provides stable local guidance, including coarse anatomical layouts and fine boundary details.

The semantic condition is introduced at the bottleneck through Frequency-Aware Attention (FA-Attention). At this stage, deep condition features from the Condition Network are combined with the flow features to facilitate interaction between the clean semantic representations and the noisy intermediate state. By modeling their relationship in a shared representation space, FA-Attention enables effective information exchange between the two streams, thereby providing high-level semantic guidance for the generation process.

MedFlowSeg is trained in an end-to-end manner.
The primary objective, $\mathcal{L}_{\mathrm{vel}}$ in Equation~\eqref{loss::vel}, measures the $L_1$ distance between the predicted velocity $\hat{u}_t$ and the constant target velocity $u_t$.
To further enhance the structural representation ability of the Condition Network, we attach an auxiliary segmentation head to its final decoder feature.
Specifically, the auxiliary loss $\mathcal{L}_{\mathrm{aux}}$ is defined as a combination of a multi-class Dice loss and a Cross-Entropy loss:
\begin{equation}
    \mathcal{L}_{\mathrm{aux}} = \mathcal{L}_{\mathrm{dice}} + \alpha \mathcal{L}_{\mathrm{ce}}.
\end{equation}

The overall training objective is then formulated by jointly optimizing the flow matching loss and the auxiliary segmentation loss:
\begin{equation}
    \mathcal{L}_{\mathrm{total}} = \mathcal{L}_{\mathrm{vel}} + \lambda \mathcal{L}_{\mathrm{aux}},
\end{equation}
where $\lambda$ and $\alpha$ are hyper-parameters that balance the auxiliary supervision and the cross-entropy term, respectively. In our implementation, we set $\lambda = 0.1$ and $\alpha = 10$.

\subsection{Structural Condition Alignment}

To provide stable structural guidance at the early stage of the flow branch, we introduce a Dual-Branch Spatial Attention (DB-SA) mechanism. Unlike directly using shallow encoder features from the condition branch, DB-SA takes the final high-resolution decoder feature of the Condition Network $f_{cond}^{-1}$ as its input.
We first extract a structural condition feature directly from this representation:
\begin{equation}
    f_{str} = f_{cond}^{-1} * k_{Conv_{3\times3}},
\end{equation}
where $k_{Conv_{3\times3}}$ is a learnable convolution kernel used to capture local spatial context.
Next, the structural condition feature is decomposed into low-frequency and high-frequency components:
\begin{equation}
\begin{gathered}
    f_{low} = \left( f_{str} * k_{Gauss}^{l} \right) * k_{Conv_{3\times3}}^{l}, \\
    f_{high} = \left( f_{str} - \left( f_{str} * k_{Gauss}^{h} \right) \right) * k_{Conv_{3\times3}}^{h},
\end{gathered}
\end{equation}
where $k_{Gauss}^{l}$ and $k_{Gauss}^{h}$ denote learnable Gaussian filters for the low- and high-frequency branches, respectively. To capture complementary structural priors at different spatial scales, the corresponding kernel sizes are set to 3 and 5. The low-frequency branch captures coarse anatomical structures, while the high-frequency branch emphasizes local boundary details.
After obtaining the complementary features, they are concatenated and converted into an adaptive attention map:
\begin{equation}
    g = \mathrm{Sigmoid}\left( \mathrm{Concat}(f_{high}, f_{low}) * k_{Conv_{1\times1}}^{g} \right),
\end{equation}
where $\mathrm{Concat}(\cdot)$ denotes channel-wise concatenation and $k_{Conv_{1\times1}}^{g}$ is a learnable $1\times1$ convolution kernel used to aggregate the fused features.
Finally, the generated attention map is applied to the first encoder feature of the Flow Network:
\begin{equation}
    f_{flow}^{\prime 0} = f_{flow}^{0} + g \cdot f_{flow}^{0},
\end{equation}
where $\cdot$ denotes element-wise multiplication. Through this process, DB-SA injects multi-frequency structural cues derived from the decoder output of the Condition Network into the shallow flow representation, thereby providing more stable anatomical guidance and sharper boundary awareness for subsequent generation.

\subsection{Semantic Condition Alignment}

The generative trajectory in flow matching evolves from noisy intermediate states to structured segmentation targets, while the Condition Network provides clean and static semantic priors extracted from the input image. This creates an intrinsic discrepancy between the two streams, particularly at the bottleneck stage where features mainly encode compressed global anatomical semantics. As the flow features are strongly timestep-dependent and often contaminated by unstable local responses during generation, directly coupling them with deterministic condition features in a shared representation space may lead to inconsistent semantic alignment and unreliable feature interaction.

To address this issue, we introduce Frequency-Aware Attention (FA-Attention), which refines the bottleneck condition feature $c^0$ using the current flow feature $f$ as a dynamic reference. As illustrated in Figure~\ref{framework}(b), FA-Attention processes $f$ and $c^0$ through two complementary pathways: a spectral branch for global semantic alignment and a spatial-token branch for discrepancy modeling.

In the spectral branch, both bottleneck features are first transformed by Fast Fourier Transform (FFT). The resulting frequency-domain features are then processed by FiLM modulation, patch embedding, and positional encoding to form token sequences $\hat{T}_f$ and $\hat{T}_c$. We perform cross-attention with $\hat{T}_c$ as the query and $\hat{T}_f$ as the key-value pairs:
\begin{equation}
    \hat{T}_F = \operatorname{CrossAttn}(\hat{T}_c, \hat{T}_f).
\end{equation}
Since FFT represents features from a global spectral view, this branch helps the condition tokens capture long-range dependencies and align coarse anatomical structures with current noisy flow state.

\begin{figure}[t]
    \centering
    \includegraphics[width=0.85\linewidth]{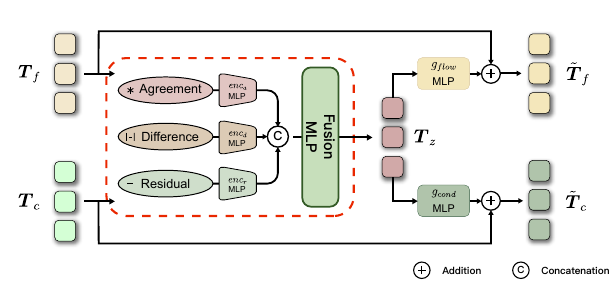}
    \caption{Architecture of the TD-X module. Given the patchified flow token $T_f$ and condition token $T_c$, TD-X constructs three complementary discrepancy cues, namely the agreement token, the difference token, and the residual token. These cues are independently encoded and fused into a unified evidence token $T_z$, which is further used to recalibrate the original tokens, yielding $\hat{T}_f$ and $\hat{T}_c$.}
    \label{tdx}
\end{figure}

In parallel, the spatial-token branch operates on the original real-valued bottleneck features, preserving the relative spatial organization of feature tokens. After FiLM modulation and patch embedding, we obtain token representations $T_f$ and $T_c$. To explicitly characterize their local mismatch, we introduce the Token Discrepancy Extractor (TD-X), as shown in Figure~\ref{tdx}. TD-X constructs three complementary cues:
\begin{equation}
    T_a = T_f \odot T_c, \quad
    T_d = |T_f - T_c|, \quad
    T_r = T_f - T_c,
\end{equation}
where $\odot$ denotes element-wise multiplication. The agreement cue $T_a$ captures co-activated responses, the difference cue $T_d$ measures discrepancy magnitude, and the residual cue $T_r$ preserves directional variation. These cues are independently encoded and fused into an evidence token $T_z$, which recalibrates the original tokens and produces discrepancy-aware spatial tokens $\tilde{T}_f$ and $\tilde{T}_c$.

Finally, a Neural Modulator integrates the spectral and spatial-token evidence. It takes $\hat{T}_F$, $\tilde{T}_f$, and $\tilde{T}_c$ as input, and applies $R=2$ time-conditioned transformation blocks to generate a modulation mask $M^0$. The mask is applied to the frequency-aware condition tokens:
\begin{equation}
    \hat{T}_C^1 = \hat{T}_F \odot (1 + M^0).
\end{equation}
The modulated tokens are then transformed back through inverse patch embedding and Inverse FFT (IFFT), yielding the refined bottleneck condition feature $c^1$. In our implementation, FA-Attention is stacked $N=4$ times. By combining global spectral alignment with spatial-token discrepancy modeling, FA-Attention provides a more stable semantic prior for subsequent vector-field estimation.

\section{Experiments}
\label{4}

\subsection{Datasets}

To evaluate our approach, we conduct experiments across five distinct medical image segmentation datasets. The public ACDC dataset~\cite{bernard2018deep}, annotated with three cardiac structures, is utilized to verify multi-class segmentation performance on cine-MRI. Additionally, to demonstrate the model's robustness across diverse imaging modalities, we employ four other public datasets: the BraTS-2021 dataset~\citep{baid2021rsna} for brain tumor segmentation in MRI, REFUGE-2~\cite{fang2022refuge2} for optic disc and cup segmentation in fundus images, the GlaS dataset~\cite{sirinukunwattana2017gland} for gland segmentation in histology, and the CAMUS dataset~\cite{leclerc2019deep} for echocardiographic segmentation in ultrasound images. 

\subsection{Implementation Details}
\label{4.2}

All models were developed in PyTorch Lightning and trained and evaluated on four NVIDIA A100 GPUs, each with 40 GB of VRAM.
All images were uniformly resized to a resolution of $256\times256$ pixels. For grayscale medical images such as MRI and ultrasound, we used a single-channel input. For color images such as fundus and histology images, the RGB format was preserved as three-channel input.
The backbone of both the Flow Network and the Condition Network follows a four-level UNet-style encoder--decoder design, with channel dimensions set to $\{64,128,256,256\}$. Each level contains two residual blocks, and the time embedding dimension is set to 256 for the time-conditioned flow branch.
We optimized the networks end-to-end using AdamW~\cite{loshchilov2017decoupled} with a batch size of 48 and an initial learning rate of $1\times10^{-4}$.
During training, we maintained an exponential moving average (EMA) of model parameters with a decay rate of 0.9999.
At inference time, we used 50 sampling steps by default, which is much fewer than the 100 sampling steps in MedSegDiff-V2~\cite{wu2024medsegdiffv2} and FlowSDF~\cite{bogensperger2025flowsdfijcv}.
For each case, the model was run 10 times to generate an ensemble of segmentation samples, which were subsequently fused using the STAPLE algorithm~\cite{warfield2004simultaneous}.
Segmentation performance was assessed using Dice, IoU, and HD95 metrics.
Higher Dice and IoU values, together with lower HD95 values, indicate better agreement with the ground truth.

\begin{table}[t]
\centering
\small
\setlength{\tabcolsep}{3pt} 
\caption{Quantitative results on the ACDC cardiac segmentation dataset. Dice and IoU scores are reported for individual cardiac structures (RV, Myo, and LV), alongside the overall average metrics including Dice, IoU, and HD95. $\uparrow$/$\downarrow$ indicates that higher/lower values represent better performance.}
\label{tab:acdc}
\begin{tabular}{cc ccc cc cc cc} 
\toprule
\multirow{2}{*}{\textbf{Category}} &
\multirow{2}{*}{\textbf{Methods}}
& \multicolumn{3}{c}{\textbf{Average}}
& \multicolumn{2}{c}{\textbf{RV}}
& \multicolumn{2}{c}{\textbf{Myo}}
& \multicolumn{2}{c}{\textbf{LV}} \\

\cmidrule(lr){3-5}
\cmidrule(lr){6-7}   
\cmidrule(lr){8-9}   
\cmidrule(lr){10-11} 

& 
& Dice $\uparrow$ & IoU $\uparrow$ & HD95 $\downarrow$
& Dice $\uparrow$ & IoU $\uparrow$
& Dice $\uparrow$ & IoU $\uparrow$
& Dice $\uparrow$ & IoU $\uparrow$\\

\midrule

\multirow{5}{*}{\begin{tabular}{c}Deterministic\\Segmentation\\Methods\end{tabular}}
& UNet                   & 87.55 & 78.37 & 9.58 & 87.10 & 77.20 & 80.63 & 67.58 & 94.92 & 90.33 \\
& nnUNet                 & 90.29 & 82.35 & 7.49 & 88.92 & 80.05 & 89.13 & 80.38 & 92.84 & 86.63 \\
& TransUNet              & 89.71 & 81.56 & 8.50 &  86.67 & 76.47 & 87.27 & 77.41 & 95.18 & 90.80 \\
& SwinUNet               & 88.07 & 78.94 & 9.03 & 85.77 & 75.09 & 84.42 & 73.04 & 94.03 & 88.73 \\
& MT-UNet                & 90.43 & 82.97 & 6.69 & 86.64 & 76.44 & 89.04 & 81.70 & 95.62 & 90.77 \\

\midrule

\multirow{5}{*}{\begin{tabular}{c}Generative\\Segmentation\\Methods\end{tabular}}
& EnsemDiff              & 92.76 & 86.61 & 6.25 & 94.52 & 89.61 & 89.03 & 80.23 & 94.73 & 89.99 \\
& SegDiff                & 91.80 & 84.90 & 6.84 & 91.68 & 84.64 & 89.34 & 80.73 & 94.37 & 89.34 \\
& MedSegDiff             & 91.89 & 85.12 & 7.16 & 92.55 & 86.13 & 87.92 & 78.44 & 93.19 & 87.25 \\
& MedSegDiff-V2          & 92.68 & 86.39 & 6.03 & 94.07 & 88.80 & 90.18 & 82.11 & 93.79 & 88.30 \\
& FlowSDF                & 91.43 & 84.25 & 6.38 & 91.28 & 83.95 & 88.71 & 79.71 & 94.30 & 89.21 \\

\midrule

Proposed
& \textbf{MedFlowSeg} & \textbf{93.57} & \textbf{87.73} & \textbf{5.27} & \textbf{94.68} & \textbf{89.90} & \textbf{90.32} & \textbf{82.35} & \textbf{95.72} & \textbf{90.97} \\

\bottomrule
\end{tabular}
\end{table}

\begin{figure}[t]
    \centering
    \includegraphics[width=0.96\linewidth]{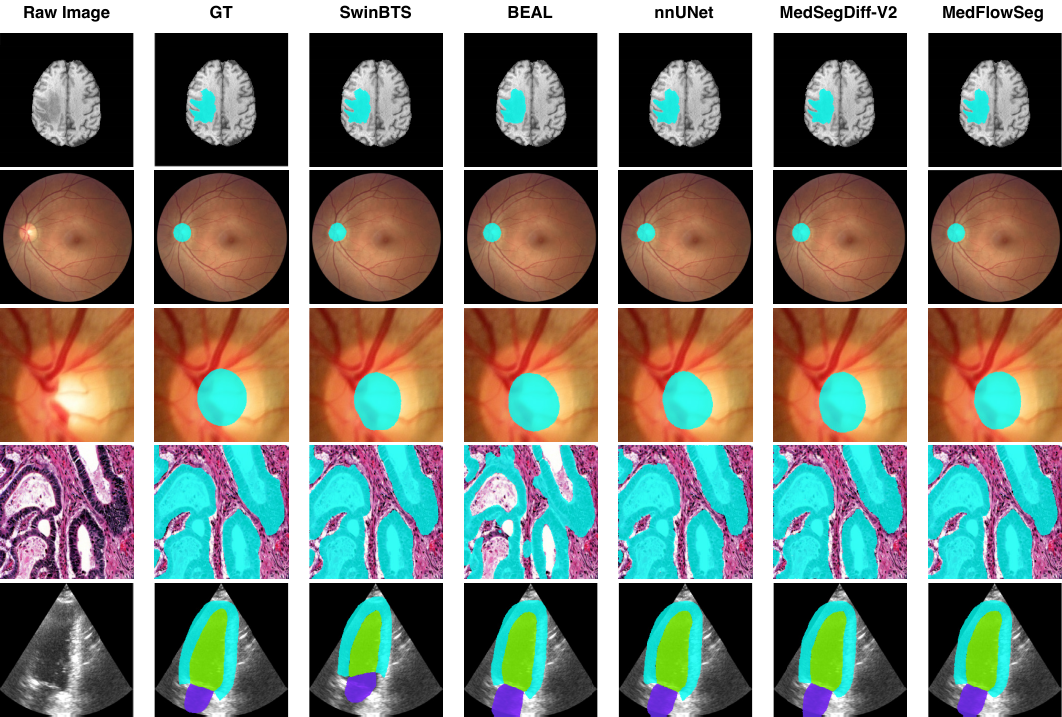}
    \caption{Visual comparison of our method against the representative baselines presented in Table~\ref{tab:multi}. The rows, from top to bottom, display brain tumor segmentation, optic disc segmentation, optic cup segmentation, gland segmentation and echocardiographic segmentation, respectively.}
    \label{vis:multi}
\end{figure}

\subsection{Comparison with SOTA Methods}

\textbf{Evaluation on multi-class medical image segmentation}
To verify the segmentation performance on cardiac structures, we evaluate MedFlowSeg on the ACDC dataset~\cite{bernard2018deep} and compare it with a wide range of representative methods, including CNN-based models (U-Net~\cite{ronneberger2015unet}, nnUNet~\cite{isensee2021nnunet}), transformer-based models (TransUNet~\cite{chen2021transunet}, Swin-Unet~\cite{cao2021swinunet}, MT-UNet~\cite{wang2022mixed}), as well as generative approaches (EnsemDiff~\cite{wolleb2022diffusion}, SegDiff~\cite{amit2021segdiff}, MedSegDiff~\cite{wu2024medsegdiff}, MedSegDiff-V2~\cite{wu2024medsegdiffv2}) and the flow-based method FlowSDF~\cite{bogensperger2025flowsdfijcv}. The quantitative results are shown in Table~\ref{tab:acdc}.

From Table~\ref{tab:acdc}, we observe that segmentation performance is not solely determined by architectural complexity or the generative paradigm. Transformer-based models (e.g., Swin-Unet) do not consistently outperform strong CNN baselines like nnUNet, and early diffusion methods (e.g., EnsemDiff, SegDiff) also fail to reliably surpass deterministic approaches. Although more advanced designs such as MedSegDiff-V2 improve performance, results remain sensitive to how conditional information is integrated.
In contrast, MedFlowSeg achieves consistent improvements across all cardiac structures, performing well on both large and thin regions. This indicates that effective conditional modeling, rather than architecture alone, is key to achieving accurate and robust segmentation.


\begin{table*}[t]
\centering
\small
\setlength{\tabcolsep}{2pt}
\caption{Comparison of MedFlowSeg against SOTA segmentation models across multiple image modalities. Methods explicitly proposed for a given task are highlighted with a grey background.}
\label{tab:multi}

\begin{tabular}{@{}cc ccccccccccc@{}}
\toprule

\multirow{2}{*}{\textbf{Task}} & \multirow{2}{*}{\textbf{Methods}} 
& \multicolumn{3}{c}{\textbf{BraTs-2021}} 
& \multicolumn{2}{c}{\textbf{REFUGE2-Disc}} 
& \multicolumn{2}{c}{\textbf{REFUGE2-Cup}} 
& \multicolumn{2}{c}{\textbf{GlaS}} 
& \multicolumn{2}{c}{\textbf{CAMUS}} \\

\cmidrule(lr){3-5} \cmidrule(lr){6-7} \cmidrule(lr){8-9} \cmidrule(lr){10-11} \cmidrule(lr){12-13}

& 
& Dice & IoU & HD95
& Dice & IoU 
& Dice & IoU 
& Dice & IoU 
& Dice & IoU \\

\midrule

\multirow{2}{*}{\begin{tabular}[c]{@{}c@{}}Brain\\Tumor\end{tabular}}
& TransBTS & \cellcolor{gray!15}87.6 & \cellcolor{gray!15}78.4 & \cellcolor{gray!15}12.44 
& 94.1 & 87.2 
& 85.4 & 75.7 
& 86.6 & 76.4 
& 88.2 & 79.4 \\
& SwinBTS  & \cellcolor{gray!15}88.7 & \cellcolor{gray!15}81.2 & \cellcolor{gray!15}10.03 
& 95.2 & 87.7 
& 85.7 & 75.9 
& 84.3 & 72.9 
& 86.9 & 79.0 \\

\midrule

\multirow{2}{*}{\begin{tabular}[c]{@{}c@{}}Optic\\Disc/Cup\end{tabular}}
& ResUNet & 78.4 & 71.3 & 18.71 
& \cellcolor{gray!15}92.9 & \cellcolor{gray!15}85.5 
& \cellcolor{gray!15}80.1 & \cellcolor{gray!15}72.3 
& 82.1 & 69.6 
& 87.9 & 82.7 \\
& BEAL    & 78.8 & 71.7 & 18.53 
& \cellcolor{gray!15}93.7 & \cellcolor{gray!15}86.1 
& \cellcolor{gray!15}83.5 & \cellcolor{gray!15}74.1 
& 84.4 & 73.0 
& 89.1 & 82.8 \\

\midrule

\multirow{2}{*}{\begin{tabular}[c]{@{}c@{}}Histology\\Gland\end{tabular}}
& DCAN   & 81.0 & 68.1 & 13.86 
& 92.3 & 85.7 
& 82.1 & 69.6 
& \cellcolor{gray!15}91.2 & \cellcolor{gray!15}83.8 
& 85.9 & 80.3 \\
& TA-Net & 79.3 & 65.7 & 15.62 
& 93.1 & 87.1 
& 82.9 & 70.8 
& \cellcolor{gray!15}90.2 & \cellcolor{gray!15}82.2 
& 86.5 & 81.5 \\

\midrule

\multirow{2}{*}{\begin{tabular}[c]{@{}c@{}}Cardiac\\Ultrasound\end{tabular}}
& UltraUNet & 84.7 & 76.9 & 13.75 
& 91.2 & 82.3 
& 83.6 & 74.2 
& 86.7 & 76.5 
& \cellcolor{gray!15}90.8 & \cellcolor{gray!15}83.1 \\
& ECHO-SegNet     & 83.2 & 71.2 & 14.79 
& 92.9 & 86.7 
& 83.5 & 71.7
& 85.2 & 74.2 
& \cellcolor{gray!15}\textbf{91.6} & \cellcolor{gray!15}84.5 \\

\midrule

\multirow{4}{*}{Det. Seg.}
& nnUNet      & \cellcolor{gray!15} 88.5 & \cellcolor{gray!15} 80.6 & \cellcolor{gray!15} 11.20 
& \cellcolor{gray!15} 94.7 & \cellcolor{gray!15} 87.3 
& \cellcolor{gray!15} 84.9 & \cellcolor{gray!15} 75.1 
& \cellcolor{gray!15} 89.1 & \cellcolor{gray!15} 80.3 
& \cellcolor{gray!15} 89.8 & \cellcolor{gray!15} 82.4 \\
& TransUNet   & \cellcolor{gray!15} 86.6 & \cellcolor{gray!15} 79.0 & \cellcolor{gray!15} 13.74 
& \cellcolor{gray!15} 95.0 & \cellcolor{gray!15} 87.7 
& \cellcolor{gray!15} 85.6 & \cellcolor{gray!15} 75.9 
& \cellcolor{gray!15} 84.6 & \cellcolor{gray!15} 73.3 
& \cellcolor{gray!15} 86.7 & \cellcolor{gray!15} 79.2 \\
& UNetr       & \cellcolor{gray!15} 87.3 & \cellcolor{gray!15} 80.6 & \cellcolor{gray!15} 12.81 
& \cellcolor{gray!15} 94.9 & \cellcolor{gray!15} 87.5 
& \cellcolor{gray!15} 83.2 & \cellcolor{gray!15} 73.3 
& \cellcolor{gray!15} 89.0 & \cellcolor{gray!15} 80.1 
& \cellcolor{gray!15} 86.2 & \cellcolor{gray!15} 80.1 \\
& Swin-UNetr  & \cellcolor{gray!15} 88.4 & \cellcolor{gray!15} 81.8 & \cellcolor{gray!15} 11.36 
& \cellcolor{gray!15} 95.3 & \cellcolor{gray!15} 87.9 
& \cellcolor{gray!15} 84.3 & \cellcolor{gray!15} 74.5 
& \cellcolor{gray!15} 87.8 & \cellcolor{gray!15} 78.3 
& \cellcolor{gray!15} 89.9 & \cellcolor{gray!15} 80.3 \\

\midrule

\multirow{5}{*}{Gen. Seg.}
& EnsemDiff           & \cellcolor{gray!15} 88.7 & \cellcolor{gray!15} 80.9 & \cellcolor{gray!15} 10.85 
& \cellcolor{gray!15} 94.3 & \cellcolor{gray!15} 87.8 
& \cellcolor{gray!15} 84.2 & \cellcolor{gray!15} 74.4 
& \cellcolor{gray!15} 87.4 & \cellcolor{gray!15} 77.6 
& \cellcolor{gray!15} 87.1 & \cellcolor{gray!15} 80.4 \\
& SegDiff             & \cellcolor{gray!15} 85.7 & \cellcolor{gray!15} 77.0 & \cellcolor{gray!15} 14.31 
& \cellcolor{gray!15} 92.6 & \cellcolor{gray!15} 85.2 
& \cellcolor{gray!15} 82.5 & \cellcolor{gray!15} 71.9 
& \cellcolor{gray!15} 87.2 & \cellcolor{gray!15} 77.4 
& \cellcolor{gray!15} 88.0 & \cellcolor{gray!15} 80.7 \\
& MedSegDiff          & \cellcolor{gray!15} 88.9 & \cellcolor{gray!15} 81.2 & \cellcolor{gray!15} 10.41 
& \cellcolor{gray!15} 95.1 & \cellcolor{gray!15} 87.6 
& \cellcolor{gray!15} 85.9 & \cellcolor{gray!15} 76.2 
& \cellcolor{gray!15} 89.4 & \cellcolor{gray!15} 80.8 
& \cellcolor{gray!15} 87.1 & \cellcolor{gray!15} 82.5 \\
& MedSegDiff-V2        & \cellcolor{gray!15} 90.8 & \cellcolor{gray!15} 83.4 & \cellcolor{gray!15} 7.53 
& \cellcolor{gray!15} \textbf{96.7} & \cellcolor{gray!15} 88.9 
& \cellcolor{gray!15} 87.9 & \cellcolor{gray!15} \textbf{80.3} 
& \cellcolor{gray!15} 90.7 & \cellcolor{gray!15} 82.9 
& \cellcolor{gray!15} 87.6 & \cellcolor{gray!15} 83.9 \\
& FlowSDF             & \cellcolor{gray!15} 86.1 & \cellcolor{gray!15} 75.6 & \cellcolor{gray!15} 12.39 & \cellcolor{gray!15} 92.7 & \cellcolor{gray!15} 86.4 & \cellcolor{gray!15} 85.7 & \cellcolor{gray!15} 75.0 & \cellcolor{gray!15} 89.9 & \cellcolor{gray!15} 81.7 & \cellcolor{gray!15} 87.5 & \cellcolor{gray!15} 82.8 \\

\midrule

\multirow{1}{*}{Proposed}
& \textbf{MedFlowSeg} 
& \cellcolor{gray!15} \textbf{91.2}& \cellcolor{gray!15} \textbf{84.5}& \cellcolor{gray!15} \textbf{6.76} 
& \cellcolor{gray!15} 95.8 & \cellcolor{gray!15} \textbf{90.3} 
& \cellcolor{gray!15} \textbf{88.6} & \cellcolor{gray!15} 80.1 
& \cellcolor{gray!15} \textbf{92.8} & \cellcolor{gray!15} \textbf{86.8} 
& \cellcolor{gray!15} 91.0 & \cellcolor{gray!15} \textbf{84.7} \\

\bottomrule
\end{tabular}

\end{table*}

\textbf{Evaluation on multi-modality medical image segmentation}
We further evaluate MedFlowSeg on multiple segmentation tasks with diverse imaging modalities.
The results are reported in Table~\ref{tab:multi}, where we compare with task-specific models such as TransBTS~\cite{wang2021transbts}, SwinBTS~\cite{jiang2022swinbts}, ResUNet~\cite{diakogiannis2020resunet}, BEAL~\cite{wang2019boundary}, DCAN~\cite{chen2016dcan}, TA-Net~\cite{wang2022ta}, UltraUNet~\cite{leclerc2019deep}, and ECHO-SegNet~\cite{dhivya2025echo},
as well as general-purpose segmentation models like UNetr~\cite{hatamizadeh2022unetr} and Swin-UNetr~\cite{hatamizadeh2022swinunetr}.

These results show that MedFlowSeg generalizes well across diverse modalities (MRI, fundus, histology, ultrasound) despite large variations in appearance, noise, and anatomy. Task-specific methods perform strongly on their target datasets but lack cross-domain robustness, while generative models are sensitive to how conditional information is integrated, leading to inconsistent gains. In contrast, MedFlowSeg achieves stable improvements across all modalities.
This robustness stems from its dual-conditioning design: DB-SA introduces multi-frequency structural priors early on, preserving both global layouts and fine boundaries, while FA-Attention aligns noisy intermediate states with clean features in spatial and frequency domains. Together, these components improve region consistency and boundary accuracy. As shown in Figure~\ref{vis:multi}, MedFlowSeg produces more coherent segmentations, particularly in challenging regions.

\subsection{Empirical Studies}
\label{4.4}

\begin{table*}[t]
\centering

\begin{minipage}[t]{0.52\textwidth}
\centering
\small
\setlength{\tabcolsep}{1.5pt}
\caption{Ablation study on the effectiveness of each component in the MedFlowSeg pipeline.}
\label{tab:ablation}
{
\renewcommand{\arraystretch}{1.33}
\begin{tabular}{cccccccc}
\toprule
\multicolumn{2}{c}{\textbf{Str. Cond.}} & \multicolumn{2}{c}{\textbf{Sem. Cond.}} & \multicolumn{3}{c}{\textbf{Datasets (Dice \%)}} \\
\cmidrule(lr){1-2} \cmidrule(lr){3-4} \cmidrule(lr){5-7}
SA & DB-SA 
& \begin{tabular}[c]{@{}c@{}}FA-A.\\(w/o Mod.)\end{tabular} 
& \begin{tabular}[c]{@{}c@{}}Neural\\Mod.\end{tabular} 
& \begin{tabular}[c]{@{}c@{}}Brain\\Tumor\end{tabular} 
& \begin{tabular}[c]{@{}c@{}}Optic\\Cup\end{tabular} 
& \begin{tabular}[c]{@{}c@{}}Histology\\Gland\end{tabular} \\
\midrule
-- & -- & -- & -- & 87.6 & 84.2 & 87.1 \\
\checkmark & -- & -- & -- & 88.5 & 85.0 & 86.9 \\
-- & \checkmark & -- & -- & 89.0 & 86.4 & 90.3 \\
-- & \checkmark & \checkmark & -- & 90.7 & 87.9 & 91.5 \\
-- & \checkmark & \checkmark & \checkmark & \textbf{91.2} & \textbf{88.6} & \textbf{92.8} \\
\bottomrule
\end{tabular}
}
\end{minipage}
\hfill
\begin{minipage}[t]{0.47\textwidth}
\centering
\small
\setlength{\tabcolsep}{1pt}
\caption{Comparison of model complexity and generated sample uncertainty. 
Params in Millions (M), Dice in percentage (\%).
}
\label{tab:uncertainty}
{
\renewcommand{\arraystretch}{1.06}
\begin{tabular}{cccccc}
\toprule
\textbf{Model} & \textbf{Params} & \textbf{Gflops} & \textbf{CI} & \textbf{GED} & \textbf{Dice} \\
\midrule
EnsemDiff & 53 & 2203 & 76.3 & 28.9 & 84.2 \\
SegDiff & 53 & 2399 & 75.4 & 26.4 & 82.5 \\
MedSegDiff & 85 & 1770 & 77.5 & 27.9 & 85.9 \\
MedSegDiff-V2 & 137 & 983 & 82.6 & 23.5 & 87.9 \\
FlowSDF & \textbf{42} & 1997 & 72.4 & 36.5 & 85.7 \\
\midrule
bone + DB-SA & -- & -- & 81.7 & 22.4 & 85.1 \\
bone + FA-A. & -- & -- & 82.9 & \textbf{21.4} & 87.4 \\
\textbf{MedFlowSeg} & 111 & \textbf{789} & \textbf{83.2} & 21.9 & \textbf{88.6} \\
\bottomrule
\end{tabular}
}
\end{minipage}

\end{table*}

\textbf{Ablation Study}
We conduct an ablation study to verify the effectiveness of each component in the dual-conditioning framework, as summarized in Table~\ref{tab:ablation}. In the table, Str. Cond. and Sem. Cond. denote structural condition and semantic condition, respectively. Starting from the backbone without the proposed conditioning modules, adding standard Spatial Attention (SA) brings only marginal improvement, indicating its limited ability to handle noisy flow features. In contrast, the proposed DB-SA consistently improves performance across all datasets by introducing multi-frequency structural priors into shallow flow features. For semantic conditioning, FA-Attention further enhances the results by reducing the discrepancy between noisy flow states and clean condition features. Finally, incorporating the Neural Modulator achieves the best performance.
\textbf{Analysis of Uncertainty}
In Table~\ref{tab:uncertainty}, we evaluate uncertainty on REFUGE2-Cup using GED and CI, measuring distribution alignment and prediction stability, respectively.
From the table, we can see that diffusion-based methods (e.g., MedSegDiff-V2) achieve relatively high diversity but suffer from larger GED. Although these methods can generate diverse samples, the stochastic sampling process often introduces instability, leading to less reliable predictions.
When using individual components, we observe that introducing DB-SA or FA-Attention alone improves either alignment or stability to some extent. However, these single-module variants cannot fully balance diversity and reliability.
By combining DB-SA and FA-Attention as MedFlowSeg, the performance is significantly improved, achieving the lowest GED together with a high CI. This suggests that the generated samples are not only diverse but also largely concentrated within the uncertainty region of the targets. Their combination leads to mutual enhancement, resulting in more consistent and accurate segmentation.

\textbf{Model Efficiency and Complexity}
In Table~\ref{tab:uncertainty}, we further compare model complexity and computational cost. Gflops measure the cost of processing a single $256 \times 256$ image until convergence, defined as when the variance over the last ten samples is below 0.1\%. MedFlowSeg achieves the lowest Gflops while also delivering the best segmentation performance.
In contrast, diffusion-based methods require many sampling steps, leading to higher cost. Parameter count is also not indicative of efficiency; e.g., MedSegDiff-V2 has higher GFLOPs due to iterative sampling.
This underscores that MedFlowSeg’s efficiency stems not only from its ODE-based formulation without stochastic denoising, but also from more effective utilization of conditional information, enabling faster convergence with fewer sampling steps and a smaller ensemble of segmentation samples.




\section{Conclusion and Discussion}

In this paper, we propose MedFlowSeg, a novel generative framework that formulates medical image segmentation as a conditional flow matching problem.To bridge the gap between noisy flow states and clean priors, we adopt a two-stream conditioning architecture with DB-SA for spatial guidance and FA-Attention for spectral alignment. Comprehensive experiments across five diverse medical imaging datasets demonstrate that MedFlowSeg significantly outperforms existing state-of-the-art models. By shifting the generative paradigm from multi-step diffusion to highly efficient, flexible flow matching, we believe MedFlowSeg provides a highly accurate and scalable solution that will serve as a strong foundation for future generative medical image analysis.

\bibliographystyle{abbrvnat}
\bibliography{refs}

\end{document}